\documentclass[10pt,twocolumn,letterpaper]{article}

\usepackage{iccv}
\usepackage[accsupp]{axessibility}
\usepackage{times}
\usepackage{epsfig}
\usepackage{graphicx}
\usepackage{amsmath}
\usepackage{amssymb}
\usepackage{float}
\usepackage{engord}
\usepackage{xcolor}
\usepackage{booktabs}

\usepackage[pagebackref=true,breaklinks=true,letterpaper=true,colorlinks,bookmarks=false]{hyperref}

\iccvfinalcopy 

\def\iccvPaperID{9573} 

\ificcvfinal\fi

\begin{document}


\title{NaviNeRF: NeRF-based 3D Representation Disentanglement 
	\\by Latent Semantic Navigation}

\author{Baao Xie\textsuperscript{1,2,3} ~Bohan Li\textsuperscript{2,4} ~Zequn Zhang\textsuperscript{2,5} ~Junting Dong\textsuperscript{6} ~Xin Jin\textsuperscript{2,3,*} ~Jingyu Yang\textsuperscript{1} ~Wenjun Zeng\textsuperscript{2,3}\\
{$^{1}$}Tianjin University ~{$^{2}$}Eastern Institute of Technology, Ningbo ~{$^{3}$}Ningbo Institute of Digital Twin ~\\{$^{4}$}Shanghai Jiao Tong University ~{$^{5}$}Northwest Normal University ~{$^{6}$}Zhejiang University\\
\vspace{-30pt}
}

\twocolumn[{%

\maketitle
\vspace{-.5mm}
\begin{figure}[H]
\hsize=\textwidth %
\centering
\includegraphics[width=.9\textwidth]{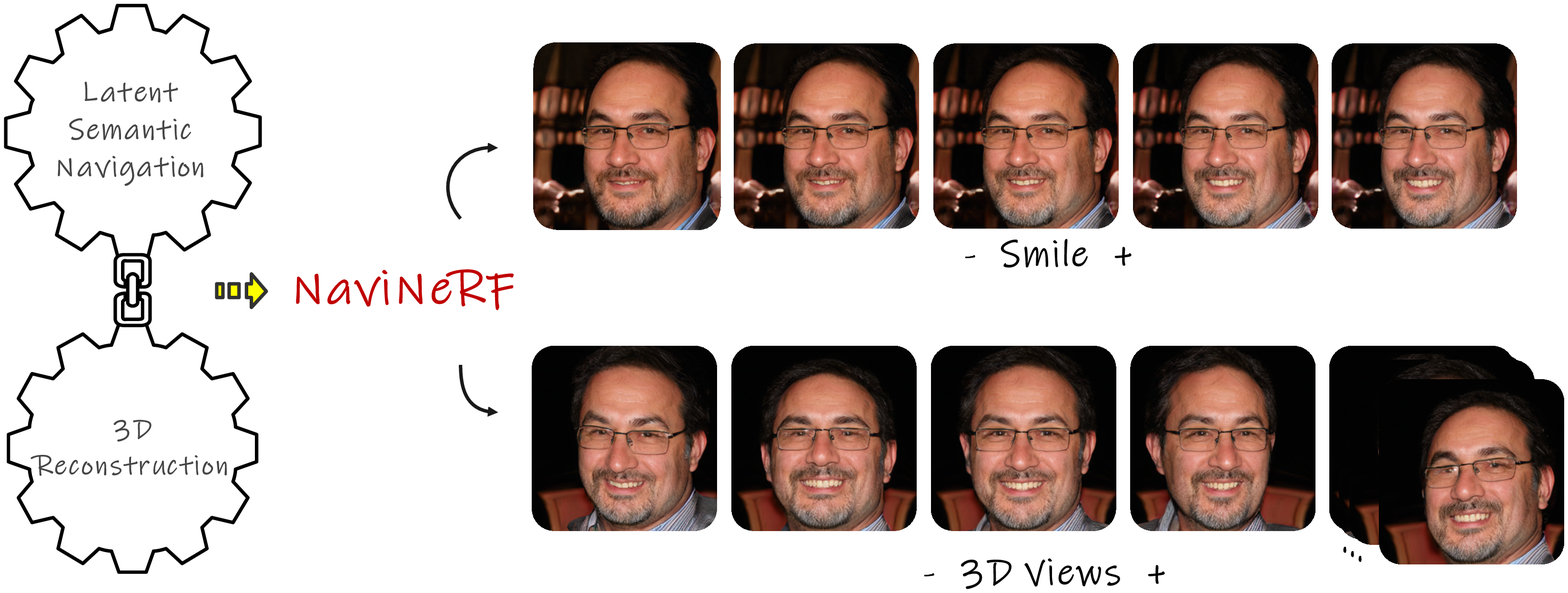}
\vspace{0.2cm}
\caption{Generated 3D objects by NaviNeRF -- a model aims to achieve fine-grained 3D disentanglement by bridging 3D reconstruction and latent semantic manipulation. The top row presents the results of shifting along the learned semantic direction that represents continuous changes in a man's mouth, visually looks like a ``smile" expression. The bottom row showcases the results of multi-view generation, which demonstrates that the attribute manipulation could still remain consistent across different views.}
\label{fig1}
\vspace{-.5mm}
\end{figure}

}]
\ificcvfinal\thispagestyle{empty}\fi
\renewcommand{\thefootnote}{*}
\begin{abstract}   
\vspace{-2mm}
3D representation disentanglement aims to identify, decompose, and manipulate the underlying explanatory factors of 3D data, which helps AI fundamentally understand our 3D world. This task is currently under-explored and poses great challenges: (i) the 3D representations are complex and in general contains much more information than 2D image; (ii) many 3D representations are not well suited for gradient-based optimization, let alone disentanglement. To address these challenges, we use \textbf{NeRF} as a differentiable 3D representation, and introduce a self-supervised \textbf{Navigation} to identify interpretable semantic directions in the latent space. To our best knowledge, this novel method, dubbed \textbf{NaviNeRF}, is the first work to achieve fine-grained 3D disentanglement without any priors or supervisions. Specifically, NaviNeRF is built upon the generative NeRF pipeline, and equipped with an \textbf{Outer Navigation Branch} and an \textbf{Inner Refinement Branch}. They are complementary —— the outer navigation is to identify global-view semantic directions, and the inner refinement dedicates to fine-grained attributes. A synergistic loss is further devised to coordinate two branches. Extensive experiments demonstrate that NaviNeRF has a superior fine-grained 3D disentanglement ability than the previous 3D-aware models. Its performance is also comparable to editing-oriented models relying on semantic or geometry priors.\footnote{Denotes the corresponding author. Code is available at \href{https://github.com/Arlo0o/NaviNeRF}{this link}.}
\end{abstract}

\section{Introduction}

3D reconstruction aims to create a virtual representation of an object or scene based on point cloud, voxel, 3D mesh, and etc. Despite significant progress of explicit reconstruction technologies such as Structure from Motion (SfM)~\cite{SFM2016}, Multi-View Stereo (MVS)~\cite{mvs2015} and Structured Light (SL)~\cite{SL2021}, it remains a critical problem that the reconstructed scenes typically lack interpretability and controllability. Thus, it is important to study the 3D representation disentanglement, in which we can identify, decompose, and manipulate the underlying explanatory factors hidden in the observed 3D data. 

However, 3D representation disentanglement is currently under-explored and faces great challenges: On one hand, the 3D representations are complex with the prohibitive storage costs, which in general contains much more information than 2D image, like depth, viewpoint, etc. On the other hand, many high-dimensional 3D representations (e.g., discrete point cloud, mesh, voxel) are essentially not well suited for gradient-based optimization~\cite{nerf2021}, which further increases the difficulty of disentanglement. All in all, how to efficiently and effectively achieve fine-grained 3D disentanglement without extra auxiliary priors or supervisions urgently needs to be solved.



Recently, the development of implicit representation learning has significantly promoted 3D reconstruction w.r.t the model flexibility and generalizability~\cite{implicit2016,implicit2021,implicit2022}. As a landmark of implicit 3D reconstruction, Neural Radiance Fields (NeRF)~\cite{nerf2021} maps scenes into a multi-layer perceptron (MLP) from limited views, which results in accurate, efficient, and differentiable 3D representations. Moreover, as a deep neural model, NeRF has preliminary shown its capability w.r.t disentangled representation learning in few studies~\cite{2021codenerf, 2022generative, 2022dfa}. Typically, the conditional NeRFs~\cite{wang2022clip, jo2021cg} achieve disentanglement with pre-defined extra latent codes, which inevitably limits the diversity of decomposed attributes. On the other branch, the editing-oriented NeRFs~\cite{sun2022cgof++, sun2022ide} also achieve a controllable 3D synthesis. However, these approaches heavily relied on geometric priors and did not identify the underlying semantic representation, such specific priors largely limited the scope of practical applications.





Revisit the recent success of disentanglement in 2D image, we knew that traversing semantically meaningful directions in the Generative Adversarial Network's (GAN)~\cite{goodfellow2020generative} latent space leads to coherent variations in the generated 2D image~\cite{goetschalckx2019ganalyze,jahanian2019steerability, peebles2020hessian, wang2021geometry}. Typically, the way of smooth navigation~\cite{voynov2020unsupervised,cherepkov2021navigating} is investigated for GAN-based semantic editing in the space of the generator’s parameters. These observations indicate that the underlying explanatory properties are probably embedded in the generative latent space. 





Based on the above discussions, in this paper, \textbf{we explore to use NeRF as a differentiable 3D representation , and introduce a self-supervised navigation to identify interpretable semantic directions in the generative latent space.} We name this novel method as \textbf{NaviNeRF}. As shown in Figure \ref{fig1}, NaviNeRF achieves a fine-grained 3D disentanglement by bridging 3D reconstruction and latent semantic manipulation. When shifting along the disentangled semantic direction that represents the mouth, we obtain a group of continuously changed visual results, look like a ``smile''. In addition, the generated results of NaviNeRF could remain 3D consistency well across different views. 

\begin{figure}[t]
 \vspace{0.3cm}
		\begin{center}
			\includegraphics[width=1\linewidth]{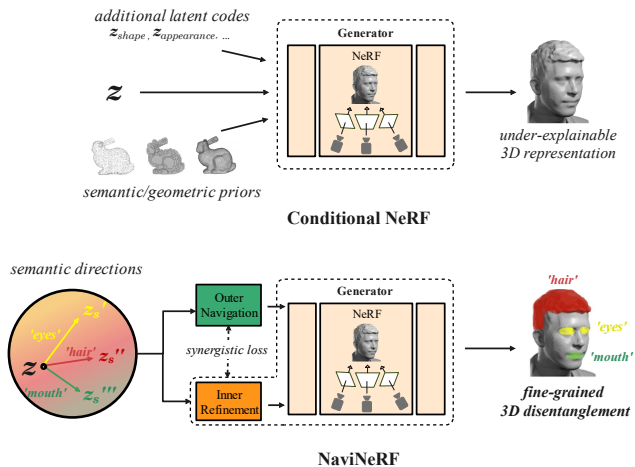}
		\end{center}
		\caption{Workflows of standard conditional NeRFs and NaviNeRF. NaviNeRF combines an outer navigation branch and an inner refinement branch by a synergistic loss, for fine-grained 3D disentanglement. Compared with existing solutions, NaviNeRF does not require conditional latent codes or semantic/geometric priors. }
		\label{fig2}
	\end{figure}

Specifically, Figure~\ref{fig2} showcases that NaviNeRF is composed of two main components: an outer navigation branch and an inner refinement branch. The outer navigation aims to identify the traversal directions as global-view factors in the latent space for disentangled representation learning —— this process employs a learnable matrix to append a shift on a latent code. The shifted code, paired with the original one, are used to generate a pair of images through the pre-trained generator. A trainable decoder is then devised to \textbf{predict the shift (i.e., semantic direction)} based on such paired images, with a reconstruction loss~\cite{voynov2020unsupervised}. Similarly, the inner branch dedicates to more fine-grained attributes by appending shifts on the \emph{specific} dimensions of intermediate latent code. Finally, a synergistic loss function is further designed to combine these two complementary branches well. Compared to off-the-shelf solutions, NaviNeRF does not resort to explicit conditional codes or any geometry priors. In summary, our contributions are:

\begin{enumerate}
    \item To our best knowledge, the proposed NaviNeRF is the first work that could achieve fine-grained 3D disentanglement at feature-level, without any priors and additional supervision.
    \item We take full advantage of both latent semantic navigation (the outer branch) and NeRF representation (the inner branch) in a complementary way. The outer branch learns to identify semantic directions for global disentangled representation learning, and the inner branch learns to focus on fine-grained attributes.
    \item As a by-product, a simple synergistic loss is designed to collaborate well two outer-inner branches within NaviNeRF.
\end{enumerate}

We evaluate NaviNeRF on two popular benchmarks: FFHQ~\cite{karras2019style} for the human face and AFHQ~\cite{choi2020stargan} for the animal face. NaviNeRF outperforms typical 3D-aware GANs including pi-GAN~\cite{chan2021pi}, GIRAFFE~\cite{niemeyer2021giraffe} and StyleNeRF~\cite{gu2021stylenerf} in attribute manipulation. Furthermore, the model obtains comparable performance to editing-oriented models which rely on semantic or geometric priors. Extensive ablation studies are also conducted to support our claims. 

\section{Related Works}
	
	Our work strongly relys on NeRF, GAN and theory of the latent semantic analysis. This section describes the relevant studies in these areas.

	\textbf{Neural implicit representations.} Encoding a scene into neural networks has shown great promise as a research direction. This includes, but is not limited to: parameterizing the geometry of a scene via signed distance functions or occupancy field~\cite{park2019deepsdf, xu2021generative}, encoding both geometry and appearance~\cite{nerf2021}, etc. One notable contribution in this field, is the work known as NeRF that has drawn extensive attention recently. It encodes a scene as 5D vector-valued function approximated by a MLP, where the function denotes spatial coordinates, viewing direction, color and density. This network can be repeatedly called by any of the volume rendering techniques to produce novel views~\cite{wang2021nerf, pumarola2021d}. The impressive performance of NeRF inspired follow-up works to extend it in alternative settings, such as training from unstructured images ~\cite{martin2021nerf, kaneko2022ar}, training without camera poses~\cite{meng2021gnerf, wang2021nerf, chen2022structure}, training with generative models~\cite{cai2022pix2nerf, kosiorek2021nerf}, etc. As a differentiable representation, NeRF and its variants have demonstrated strong capabilities at generating 3D scenes with high accuracy, efficiency and consistency. However, these approaches are commonly deficient in interpretable control over partial properties, such as shape, color, texture, and lighting.
	
    \textbf{3D-aware GAN models.} Recently, generative models have brought NeRF a certain degree of scene control capabilities. The early attempts in this routine are GRAF~\cite{schwarz2020graf} and pi-GAN~\cite{chan2021pi}. The former handles category-specific generation by conditioning NeRF on shape and appearance codes. Following the NeRF pipeline, the generator can synthesize an image by taking random codes and camera poses. The generated image is fed into the discriminator along with real images, thus implementing a GAN. pi-GAN is similar to GRAF, but conditions on a single latent code and utilizes FiLM-SIREN layers~\cite{sitzmann2020implicit, dumoulin2018feature} instead of simple MLPs. Encoding additional latent codes enriches the model with disentangled capability, but double-edgedly, limits the range of disentangled attributes. 
    
    Another impressive work in this domain is GIRAFFE~\cite{niemeyer2021giraffe}, which represents scenes as compositional generative NeRFs without any additional supervision. It is in capacity of disentanglement on separating background and foreground of the scene. Although the compositional architecture delivers control capacity over object-level, individual's local attributes are not yet fully disentangled. More recently, ~\cite{gu2021stylenerf} integrates NeRF into a StyleGAN~\cite{karras2020analyzing} based generator to produce high-resolution and multi-view consistent 3D scenes. As a by-product, it inherits the style control ability from StyleGAN baseline but still, fails to disentangle on detailed attributes.

    \textbf{3D representation editing methods.} Except embedding additional codes into latent space, many editing-oriented NeRFs adopt option to enhance 3D perception by leveraging semantic or geometric priors~\cite{sun2022cgof++, hong2022headnerf, kania2022conerf}. With GAN inversion technology~\cite{jahanian2019steerability, zhu2016generative, huh2020transforming}, these approaches can edit specific regions of a 3D scene under interactive controls given by user. Typically, ~\cite{sun2022fenerf} trains the model using paired monocular images and semantic maps, and obtains locally-editable images. Furthermore, ~\cite{yuan2022nerf} extracts an explicit triangular mesh representation as geometric priors, which can then be intuitively deformed by the user for 3D editing. Although these methods could obtain promising results on pixel-wise editing, they are still inapplicable to perform interpretable disentanglement since the underlying semantic representations are not essentially learned. 

    \textbf{Latent semantic analysis.} Empirical studies have revealed that GAN latent spaces are embedded with interpretable semantic information ~\cite{goetschalckx2019ganalyze, plumerault2020controlling, shen2021closed, khrulkov2021disentangled}. Representatively, ~\cite{shen2020interpreting} verifies that GANs trained with face images have latent spaces that contain semantic directions corresponding to specific facial features. Since such interpretable directions provide a straightforward route to robust image editing, their discovery currently receives much research attention. ~\cite{shen2021closed} further proposes the classifiers pre-trained on the facial data, to predict certain face attributes. These classifiers are then used to produce pseudo-labels for the generated images and their latent codes. Furthermore, ~\cite{jahanian2019steerability} conducts series of experiments to verify that the interpretable directions are responsible for diverse specific features, by maximizing the score of pre-trained generative models. The aforementioned semantic analysis could potentially enrich the exploration of latent space in generative NeRFs, which therefore resolve the defectiveness of semantic information in 3D scenarios. Motivated by this idea, our model is proposed to complementally take advantage of NeRF representation and semantic direction manipulation, targeting on fine-grained 3D disentanglement.
    
\begin{figure}[t]
 \vspace{0.3cm}
    \begin{center}
        \includegraphics[width=.7\linewidth]{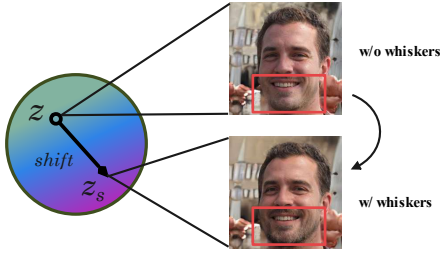}
    \end{center}
    \caption{Within a latent space $\mathcal{Z}$, the model is proposed to discover and manipulate a interpretable semantic direction from original code $z$ to shifted code $z_s$. Traversing along this direction lead to continue changes on a disentangled representation of generated image.}
    \label{fig3}
\end{figure}

\section{Methodology}
\begin{figure*}
   \vspace{0.3cm}
	\centering
	\includegraphics[width=1\textwidth]{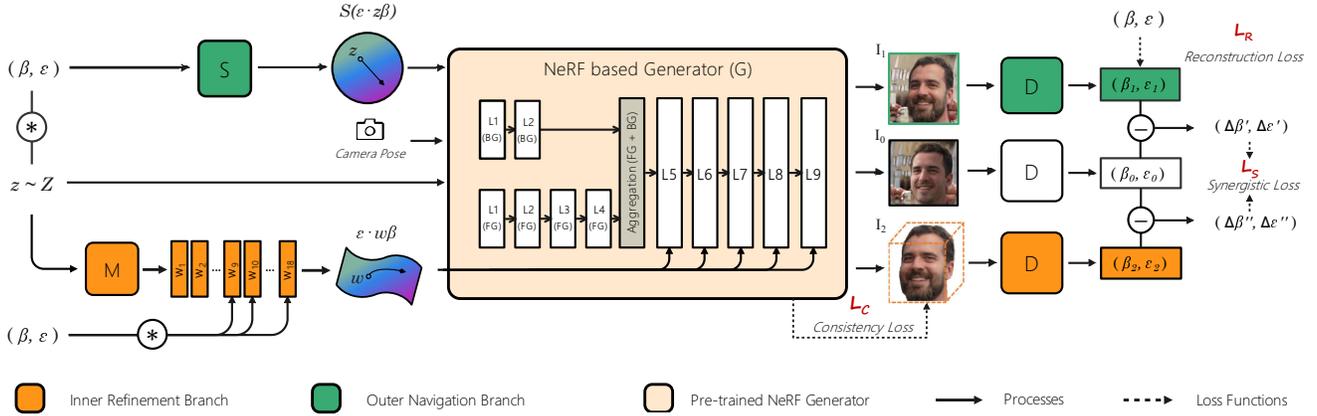}
	\caption{NaviNeRF is characterized by two complementary branches, termed as the outer navigation branch and the inner refinement branch. The former, depicted in green, appends a shift on sampled latent code $z$ through a learnable matrix $S$. $z$ and shifted code $z_s$ are used to generate paired images, which are devoted to train the decoder $D$ for semantic direction identification. The latter, shown in orange, produces fine-grained awareness and 3D consistency by appending shifts on specific dimensions of intermediate latent vector $w$. The two branches are combined by a synergistic loss, ultimately achieving feature-level 3D disentanglement.}
	\label{fig4}
\end{figure*}

In this section, we elaborate the three key modules that constitute our model: an outer navigation branch, an inner refinement branch and a synergistic loss for interaction. The architecture of the model is illustrated in Figure \ref{fig4}. Correspondingly, we start the introduction of the navigation branch in Section 3.1, and elaborate the structure of the refinement branch in Section 3.2. Then in Section 3.3, the synergistic loss coordinating the two branches are described, along with other loss functions we employed. 

\subsection{Outer Navigation Branch}
The navigation branch, inspired by previous studies of latent semantic manipulation~\cite{voynov2020unsupervised, cherepkov2021navigating, ren2021learning}, is proposed to identify interpretable semantic directions in the latent space. Specifically, given a pre-trained GAN generator $G$, which maps the samples $z \sim N(0,\mathbb{I}) $ drawn from the latent space $\mathcal{Z}$ to generate multi-view images. Our objective is to learn a set of semantic directions, that facilitate continuous changes of specific attributes in the generated image upon being shifted along each direction. For example, Figure \ref{fig3} visualizes the alter on the object's whiskers in RGB space, which aroused by the traverse from $z$ to $z_s$ in latent space. 

Towards this target, we propose a learnable matrix $S\in \mathbb{R}^{d}$ that enables the identification of a shift with a specific direction index $\beta$ and scale $\epsilon$. Here, $d$ represents the dimensionality of the matrix, where the columns of $S$ corresponding to candidate directions. The shift $S$ is appended on $z$ for conducting a shifted code $z_s$. Both $z$ and $z_s$ are then input to the pre-trained $G$ to produce paired images $I_0$ and $I_1$ respectively: 

$$ I_0 = G(z), I_1 = G(S(\epsilon\cdot z \beta)) $$
where the diversity between the two images being solely attributed to this latent shift. In other words, $I_1$ is the transformation of $I_0$, corresponding to moving by $\epsilon$ along the direction $\beta$.

After the reconstruction, a trainable decoder $D$ is proposed to project generated images from RGB space back to latent space. More specifically, $D$ is a function that maps images pairs into the shift increment, parameterized by a MLP network. The shift increment ($\Delta \beta '$, $\Delta \epsilon'$) from $I_0$ to $I_1$, is constrained with the ground truth by the reconstruction loss $\mathcal{L_R}$.

Although the outer module is devised to discover semantic directions, it alone is insufficient for 3D scenarios owing to the deficiency of geometric consistency. Nevertheless, additional instructions are expected, for model to concentrate on fine-grained attributes over discovered directions. Towards it, we propose an inner refinement branch for two goals: obtaining the perception of fine-grained representations and, preserving 3D consistency.

\subsection{Inner Refinement Branch}
Within the inner branch, we adopt StyleNeRF baseline as the generator, which takes integrated advantages of NeRF and StyleGAN. In a basic GAN, latent code $z$ is sampled directly from a Gaussian distribution and determines the global style of the generated image. However, the limited capacity of the normal distribution constraints the disentanglement capability of $\mathcal{Z}$ ~\cite{xia2022gan}. Differently, StyleGAN maps native $z$ to a layer-wise style code $w$ by a 8-layer mapping network $M$. The intermediate latent space is referred as $\mathcal{W^+}$ space that contains more disentangled features than $\mathcal{Z}$. Magnetized by the disentanglement capability of $\mathcal{W^+}$, we devise to build paired codes (original and shifted) similar as the outer branch, but with the shifts over $w$.

Tentatively, semantic shifts are appended on each dimension of latent code $w$, which however, leads to an unexpected entanglement among global style and fine-grained details. To our knowledge, the phenomenon is attributed to the different control scope of distinct dimensions in $w$. Specifically, it has been observed that dimensions of $w$ correspond to different levels of details, roughly in three groups: global, coarse, and fine~\cite{karras2020analyzing, richardson2021encoding, patashnik2021styleclip, zheng2022sdf}. Conditioning partial details together with global and coarse style can pose a challenge for disentanglement. In this sense, we turn to align the shifts on \engordnumber{9} - \engordnumber{18} dimensions of $w$ which theoretically, controls the fine-grained attributes, therefore encouraging the model to learn partial representations.

Following the architecture of StyleNeRF, we perform NeRF++ behaving as the NeRF synthesis network. It comprises a foreground (FG) NeRF in a unit sphere and a background (BG) NeRF represented using an inverted sphere parameterization~\cite{zhang2020nerf++}. As shown in Figure \ref{fig4}, two MLPs are utilized separately to predict the density. A shared MLP is then employed with up-sample blocks for color prediction. 

To achieve disentangled semantic manipulation, the shifted dimensions (\engordnumber{9} - \engordnumber{18}) of $w$ are two-to-one fed into \engordnumber{5} - \engordnumber{9} NeRF MLP layers through an affine transformation, followed by a decoder to predict shift increment ($\Delta \beta ''$, $\Delta \epsilon''$).

\subsection{Loss Functions}
The heart of the NaviNeRF is at the complementarity of navigation and refinement branches. To combine these two modules, a synergistic loss is devised in three steps: (i) decoding generated image pairs into the shift with the direction index $\beta$ and scale $\epsilon$; (ii) calculating the increments of both branches, termed as ($\Delta \beta '$, $\Delta \epsilon'$) and ($\Delta \beta ''$, $\Delta \epsilon''$) respectively; (iii) minimizing the distance of two increments. The variation of the two modules can be calculated as:

$$ (\Delta \beta ', \Delta \epsilon') = D(G(z)), G(S(\epsilon\cdot z \beta)))$$
$$ (\Delta \beta '', \Delta \epsilon'') = D(G(z)), G(\epsilon\cdot w \beta))$$

Generally, the synergistic loss is proposed to be a cross-entropy loss, which therefore can be demonstrated as:

$$ \mathcal{L_S}  = CrossEntropy((\Delta \beta ', \Delta \epsilon'), (\Delta \beta '',  \Delta \epsilon''))$$

In addition to the synergistic loss, we utilize the reconstruction loss $\mathcal{L_R}$ on generated images and ground truth to optimize the reconstruction quality of outer branch using a MSE loss.

$$ \mathcal{L_R} = MSE((\Delta \beta ', \Delta \epsilon'), (\beta, \epsilon))$$

We also apply the consistency loss $\mathcal{L_C}$ to enforce 3D consistency, as instituted by~\cite{gu2021stylenerf}. More formally, we propose another original NeRF path without up-sampling blocks in the NeRF generator, for producing a low-resolution but consistent image to supervise $I_2$ by the consistency loss. In this way, $I_2$ can be closer to the NeRF results, which have multi-view consistency. The loss is calculated as follows:
$$ \mathcal{L_C} = \frac{1}{\left | P \right | } {\textstyle \sum_{(i,j)\in P}^{}} (I_2[i,j]-I_{NeRF}[i,j]))^2$$
where $P$ denotes randomly sampled pixels. $I_2$ is instantiated as the image generated from the inner branch and $I_{NeRF}$ is the image from the original NeRF.

In such setting, the total loss can be summarized as:

$$ \mathcal{L}_{total} = \mathcal{L_S} + \lambda_R \mathcal{L_R} + \lambda_C \mathcal{L_C}$$ 
where the $ \lambda_R $ and $ \lambda_C $ are the hyper-parameter. In default, we adopt $ \lambda_R $ = 0.8 and $ \lambda_C $ = 0.6 to balance the disentanglement capability and reconstruction quality.

\begin{figure*}
	\centering
		\includegraphics[width=1\linewidth]{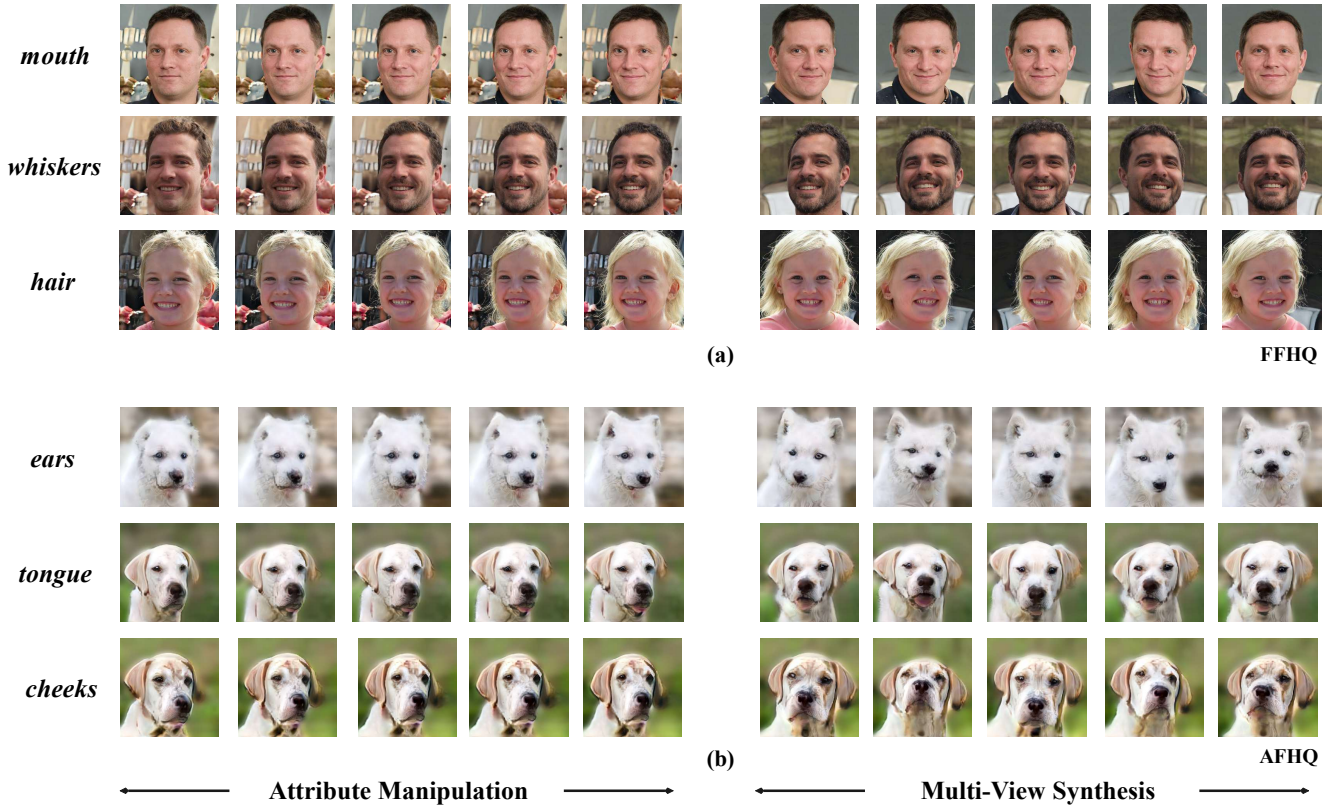}
	\caption{Fine-grained 3D Disentanglement Results of NaviNeRF. The left columns present the results of attribute manipulation and the right columns showcase corresponding 3D reconstruction results. Respectively, (a) demonstrates the semantic manipulation on the FFHQ dataset including the man's mouth, whiskers and the girl's hair; (b) shows the manipulation results on the puppy’s ears, tongue and cheeks.}
	\label{fig5}
\end{figure*}

\section{Experiments}
\subsection{Experimental Settings}
We conduct a set of experiments on several datasets: \textbf{FFHQ~\cite{karras2019style}} consists of 70,000 high-quality images of human faces; \textbf{AFHQ~\cite{choi2020stargan}} contains 15,000 high-quality images at a resolution of in three categories of cat, dog, and wildlife; \textbf{CompCars~\cite{yang2015large}} contains 136726 images capturing the entire cars with different styles; \textbf{LSUN~\cite{yu2015lsun}} consists of about one million images for multiple object categories. We pre-train the generator with resized images from aforementioned datasets at 256$\times$256 resolution for a trade-off on quality and controllability. For pre-training, we follow the instructions outlined in StyleNeRF adopting batch size in 64 and a learning rate as 0.0025. To train outer and inner branches, we apply batch sizes of 64 and 32 for FFHQ and AFHQ, respectively, with a learning rate of 0.0005. All experiments were performed on 4 Nvidia GPUs (Tesla A100 80GB) with CUDA version 11.6. 

\subsection{Results}
\subsubsection{Fine-grained 3D Disentanglement}
We firstly conduct experiments on two popular datasets: FFHQ and AFHQ, in aim to evaluate the effectiveness of our proposed model in feature-level 3D disentanglement. These benchmarks consist of high-resolution images of single objects and cover a diverse range of real-world scenarios, highlighting the robustness of our results. We further expand the model's applicability to more intricate scenes by leveraging CompCars and LSUN datasets, which effectively showcasing the model's capacity for generalization.

In Figure \ref{fig5}, we present several visual effects on different objects induced by the discovered directions, paired with their reconstruction results. More specifically, we illustrate the attribute manipulation results on the left side. And the results of the 3D reconstruction are demonstrated on the right, visually formed as the multiple perspectives of the image after manipulation (the \engordnumber{5} image on the left). It is worth mentioning that the quality of the reconstructed images is highly dependent on the pre-trained generator. But even the model is fully configured as StyleNeRF in the pre-training stage, artifacts still exist in generated images for the AFHQ dataset. However, with such ambiguous and challenging data, the model still achieves fine-grained awareness, which validates its robustness.

Different from those pixel-level editing works, our model achieves semantic manipulation in latent space, which is easily extendable to more scenes per the theory of disentangled representation learning~\cite{bengio2013representation}. Moreover, the differentiable and continuous nature of the NeRF-based implicit representation makes it more flexible and geometry-free for novel view synthesis. Furthermore, NaviNeRF also inherits the well-studied properties of StyleGAN, leading to a universal feature generation capability. To emphasize the generalization capability of the model, we showcase extra fine-grained disentanglement results within more generic scenes in Figure \ref{figLSUN}. Supplementary examples can be found in the appendix and in the repository.       

\subsubsection{Qualitative Comparison}

\textbf{NaviNeRF vs. typical 3D-aware GANs.} Typical 3D-aware GAN such as pi-GAN, GIRAFFE and StyleNeRF claim a certain degree of disentanglement in their model. Thereinto, pi-GAN and GIRAFFE provide scene control over object appearance, style, and rotation by altering shape code $z_{shape}$ and appearance code $z_{app}$. On the other hand, StyleNeRF allows for the manipulation of global styles by leveraging the disentanglement capacity of StyleGAN. Since GIRAFFE and StyleNeRF provide pre-trained models on the FFHQ, we directly load their checkpoints and re-trained pi-GAN with the same data. However, as these models control the scenes in different ways, comparing and visualizing the results of 3D semantic manipulation can be a challenging task. To ensure fairness, we adopt style interpolation for each typical GAN over the same attribute to demonstrate their capability of fine-grained control. Specifically as shown in Figure \ref{fig6}, we manually select two images to determine an assumed direction for these 3D-aware GANs without the function of direction manipulation. In the first image, the mouth of the object is closed while in the second image, the mouth is smiling. The other regions of the face in both images are identical. We then extract the paired latent code $z$ and $z_s$ corresponding to these two images. By further manipulating along the direction from $z$ to $z_s$, we can thus conduct specific attribute (mouth) control results for each model.
\begin{figure}[t]
	\centering
	\includegraphics[width=1\linewidth]{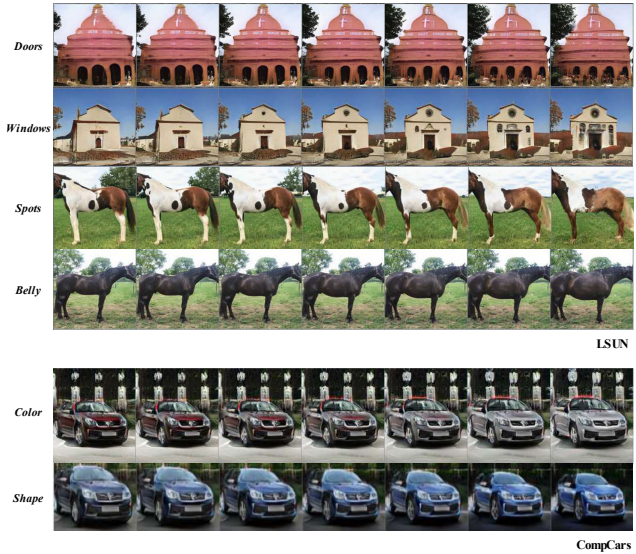}
	\caption{Fine-grained 3D disentanglement results on LSUN and CompCars datasets that covering more general scenes, to demonstrate both the disentanglement and generalization capability of the model. }
	\label{figLSUN}
\end{figure}
\begin{figure}[t]
    \begin{center}
        \includegraphics[width=1\linewidth]{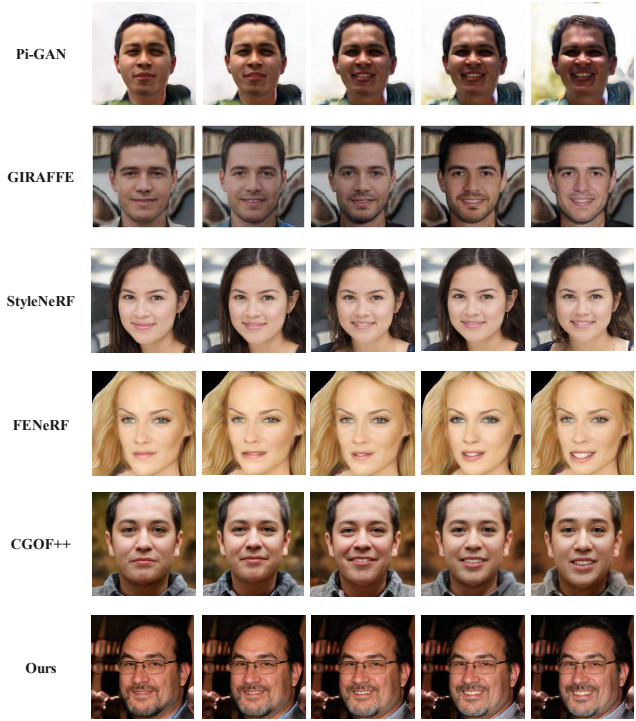}
    \end{center}
    \caption{Comparison results of NaviNeRF with typical 3D-aware GANs and 3D editing-oriented models on the FFHQ dataset. The first three rows present disentanglement results on manipulating object's mouth from pi-GAN, GIRAFFE and StyleNeRF, respectively. The \engordnumber{4} - \engordnumber{5} rows demonstrate the manipulation results of two editing-oriented models.}
    \label{fig6}
\end{figure}

In the first three rows of Figure \ref{fig6}, we represent the comparison results of continue interpolation upon the man's mouth. The results obtained by pi-GAN and GIRAFFE show that global styles such as beard, skin tone and eye socket are changing simultaneously during the manipulation on the mouth. Although StyleNeRF produces better results, some partial representations remain entangled such as hairstyle. NaviNeRF demonstrates better overall performance on disentangling specific attributes compared to other baselines, which confirms the superiority of the model for feature-level 3D disentanglement. 
 
~\\
\noindent \textbf{NaviNeRF vs. 3D editing-oriented models.}
Alternately, another bunch of work tends to attain 3D scene control by incorporating semantic or geometric priors. Encoded priors can enrich the model 3D perception and enable editing of specific attributes through inversion techniques. Accordingly, we compare NaviNeRF with two novel works reported in 2022 within this domain: FENeRF~\cite{sun2022fenerf} and CGOF++~\cite{sun2022cgof++}. The former learns 3D representation from widely available monocular images and semantic mask pairs. The model uses the semantic mask to manipulate partial attributes via GAN inversion. The latter is a conditional NeRF that incorporates a mesh-guided sampling process and a depth-aware density regularizer. For comparison, we load their pre-trained models on the FFHQ dataset and conduct the samples for editing object's mouth.

The last three rows of Figure \ref{fig6} demonstrate comparison results for FENeRF, CGOF++, and NaviNeRF. Three methods achieve approximate results on editing the partial attribute. Although the extra priors offer a shortcut for scene disentanglement, the editing-oriented models inevitably suffer from its redundancy, inflexibility and inefficiency. In addition, these models are dictated to edit a specific pixel area but not essentially understand the underlying semantic meanings. As emphasized earlier, our model aims to autonomously learn the latent semantic information, without any priors and additional supervision. Therefore, the comparable results demonstrate the superiority of our model.

\subsubsection{Quantitative Comparison} Table 1 reports the results of Frechet Inception Distance (FID)~\cite{heusel2017gans} and Kernal Inception Distance (KID)~\cite{heusel2017gans} scores to measure the quality of the generated images. Our model outperforms other typical 3D-aware models and is slightly inferior to StyleNeRF. To our understanding, the decrease in generation quality compared with StyleNeRF can be a trade-off for fine-grained control (i.e.,we did not fine-tune the pre-trained generator in attribute manipulation). Furthermore, the performance decrease in the AFHQ dataset, which we believe, is caused by the non-fully trained generator as noted in Section 4.2.1.

\begin{table}[H]
  \centering
  \caption{Quantitative comparison results with typical 3D-aware models in FID and KID $\times10^3$.}
  \vspace{0.3cm}
    \begin{tabular}{ccccc}
    \toprule
    \makebox[0.08\textwidth][c]{} & \makebox[0.06\textwidth][c]{\textbf{FFHQ}} & \makebox[0.06\textwidth][c]{ }& \makebox[0.06\textwidth][c]{\textbf{AFHQ}} & \makebox[0.06\textwidth][c]{ }\tabularnewline
    Models & FID   & KID  & FID   & KID  \tabularnewline
 \midrule
    pi-GAN  & 87  & 99 & 53  & 35.4 \tabularnewline
    GIRAFFE & 38 & 25.7 & 36  & 14.7 \tabularnewline
    StyleNeRF & 10.4 & 4.6 & 16  & 4.3 \tabularnewline
\midrule
    \textbf{Ours} & 13 & 6.9 & 22  & 9.1  \tabularnewline
\bottomrule
    \end{tabular}%
  \label{table1}%
\end{table}%

\begin{figure}[t]
    \begin{center}
        \includegraphics[width=1\linewidth]{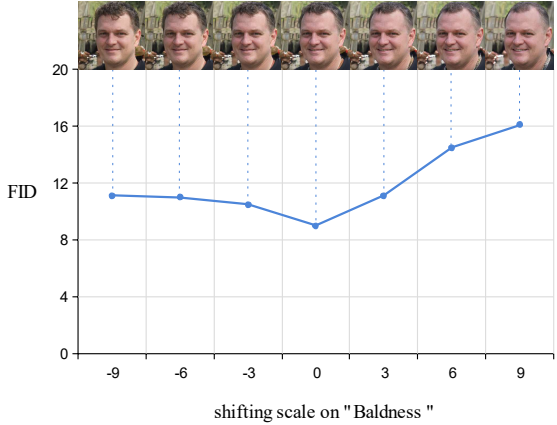}
    \end{center}
    \caption{FID scores of images with different shifting magnitudes on the man's baldness.}
    \label{fig8}
\end{figure}

Figure \ref{fig8} demonstrates the FID scores for images with various shifting magnitudes in the ``baldness" direction. The results indicate that the model can maintain the generation quality across the attribute manipulation. It supports our claims that the model has learned to manipulate fine-grained features while maintaining a consistent global style.

Table 2 and 3 presents the method budget on the FFHQ to reveal that NaviNeRF is high-efficient. Table 2 shows that two-branch design did not increase the feature size and we also use the pre-trained generator for lower cost, thus avoid high memory issue in rendering. As demonstrated in Table 3, NaviNeRF achieves significant speedup over pure NeRF-based methods and is comparable to StyleNeRF. 

\begin{table}[H]
  \centering
  \renewcommand\arraystretch{1.2}
  \caption{Training Budget report on parameters size, FLOPs, memory cost and training time (TT).}
  \vspace{0.3cm}
    \resizebox{\linewidth}{!}{\begin{tabular}{ccccc}
    \toprule
    Budget & Params(M) & FLOPs(G) & Mem(G) & TT(hrs)
    \tabularnewline
 \midrule
    Single Branch& 11.9 & 142.9 & 23 & 5.8\tabularnewline
    Double Branches& 12.2 & 162.4 & 25 & 7.5\tabularnewline
\bottomrule
    \end{tabular}}
  \label{table2}%
\end{table}%

\begin{table}[H]
 \vspace{-0.5cm}
\renewcommand{\arraystretch}{0.8}
  \centering
  \renewcommand\arraystretch{1.2}
  \caption{Rendering time (RT) comparisons with other methods at $256^2$. }
  \vspace{0.3cm}
    \resizebox{\linewidth}{!}{\begin{tabular}{cccccc}
    \toprule
    Models & pi-GAN & GIRAFFE   & StyleNeRF & \textbf{Ours}  \tabularnewline
 \midrule
    RT(ms/image) & 785 & 181 & 75 & 97 \tabularnewline
\bottomrule
    \end{tabular}}
  \label{table3}%
\end{table}%

\subsection{Ablation study}
To validate the effectiveness of some key designs in NaviNeRF, we conduct ablations over w/o synergistic loss, choice of shifting dimensions and choice of latent space. 

\begin{figure}[t]
    \begin{center}
        \includegraphics[width=1\linewidth]{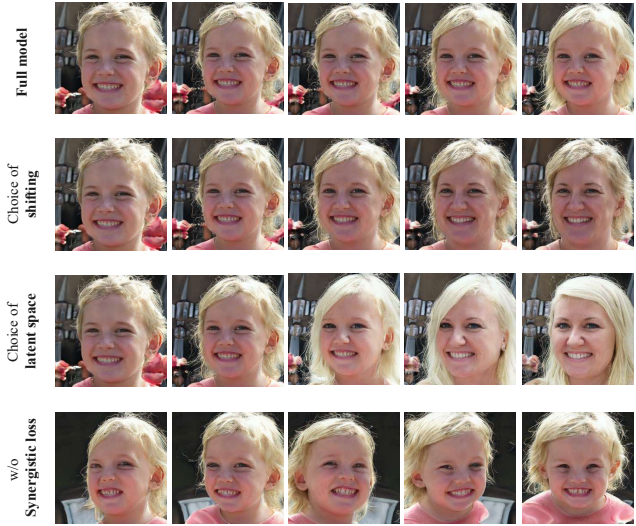}
    \end{center}
    \caption{Ablation studies on manipulating the object's ``hair" over the choice of shifting dimensions, the choice of latent space and w/o synergistic loss.}
    \label{fig7}
\end{figure}

~\\
\noindent \textbf{Shifting on $w$ dimensions.}
As previously mentioned, different $w$ dimensions correspond to three levels of partial style: \engordnumber{1} - \engordnumber{4} layers determine the global features, \engordnumber{5} - \engordnumber{8} layers for coarse features and \engordnumber{9} - \engordnumber{18} layers for fine-grained details. In Figure \ref{fig7}, we compare the reconstruction results of shifting on \engordnumber{9} - \engordnumber{18} layers (full model) against with shifting on every 18 layer (second row) in $w$. When shifting on every $w$ dimension, global features such as face shape, skin texture and wrinkle style are variously entangled, resulting in an older appearance for the girl. It indicates that shifting on specific dimensions can enforce the model to concentrate on fine-grained representations. 

~\\
\noindent \textbf{$\mathcal{W}$ space vs. $\mathcal{W^+}$ space.}
Referring to StyleNeRF, $\mathcal{W}$ is an intermediate space with a distribution matching better to the real data compared with the original $\mathcal{Z}$ space. It contains a single intermediate $w$ vector, whereas the $\mathcal{W^+}$ space comprises 18 different style vectors. To examine the impact of latent space choice, we append identical shifts on $\mathcal{W}$ and $\mathcal{W^+}$. In the case of $\mathcal{W}$ space, we duplicate the single $w$ into 18 dimensions. The semantic shifts are then appended on the \engordnumber{9} - \engordnumber{18} dimensions of the intermediate vectors from both spaces. The third row in Figure \ref{fig7} demonstrates that, shifting on $\mathcal{W}$ space enable the model to control global styles but fails to achieve fine-grained manipulation. That is why we choose to propose our model on $\mathcal{W^+}$ for feature-level 3D disentanglement.

~\\
\noindent \textbf{w/o synergistic loss.} 
We perform NaviNeRF with or without the synergistic loss to study its impact. Since the inner branch cannot be directly trained without the synergistic loss, we replace it with a reconstruction loss which is in the same configuration as $\mathcal{L_R}$. The \engordnumber{4} row of Figure \ref{fig7} demonstrates that removing the synergistic loss leads to severe 3D inconsistent artifacts during disentanglement. The result supports the notion that synergistic loss combines the outer and inner branches in a complementary way to achieve fine-grained disentanglement and 3D consistency.

\section{Conclusion}
In this paper, we present NaviNeRF, a NeRF-based 3D reconstruction model that achieves fine-grained disentanglement while preserving 3D accuracy and consistency without any priors and supervision. The model consists of two complementary branches: an outer navigation branch delicate to identify the traversal directions as factors in the latent space, while an inner refinement branch produces fine-grained awareness and 3D consistency. We also design a synergistic loss to combine the two modules. The model is evaluated on challenging datasets to demonstrate its ability of fine-grained disentanglement in 3D scenarios. The experimental results indicate that NaviNeRF outperforms typical conditional NeRFs. Furthermore, its performance is also comparable to editing-oriented models relying on semantic or geometry priors, which supports our claims.
\section*{Acknowledgement}
This work was supported in part by ZJNSFC under Grant LQ23F010008. 

{\small
\bibliographystyle{unsrt}
\bibliography{iccv23_NaviNeRF_cameraready}
}

\end{document}